\documentclass{article}

\usepackage{microtype}
\usepackage{graphicx}
\usepackage{subfigure}
\usepackage{booktabs} 
\usepackage{multicol}

\usepackage{hyperref}

\usepackage[accepted]{icml2020_arxiv}

\icmltitlerunning{Finding online neural update rules by learning to remember.}

\begin{document}

\twocolumn[
\icmltitle{Finding online neural update rules by learning to remember.}



\icmlsetsymbol{equal}{*}

\begin{icmlauthorlist}
\icmlauthor{Karol Gregor}{to}
\end{icmlauthorlist}

\icmlaffiliation{to}{DeepMind}

\icmlcorrespondingauthor{Karol Gregor}{karolg@google.com}

\icmlkeywords{Machine Learning, ICML}

\vskip 0.3in
]



\printAffiliationsAndNotice{}  

\begin{abstract}
We investigate learning of the online local update rules for neural activations (bodies) and weights (synapses) from scratch. We represent the states of each weight and activation by small vectors, and parameterize their updates using (meta-) neural networks. Different neuron types are represented by different embedding vectors which allows the same two functions to be used for all neurons. Instead of training directly for the objective using evolution or long term back-propagation, as is commonly done in similar systems, we motivate and study a different objective: That of remembering past snippets of experience. We explain how this objective relates to standard back-propagation training and other forms of learning. We train for this objective using short term back-propagation and analyze the performance as a function of both the different network types and the difficulty of the problem. We find that this analysis gives interesting insights onto what constitutes a learning rule. We also discuss how such system could form a natural substrate for addressing topics such as episodic memories, meta-learning and auxiliary objectives.
\end{abstract}

\section{Introduction}
\label{introduction}

A brain is a type of recurrent neural network, in the sense that it receives and sends information online, and consists of neurons with synapses as core computational components. This computational structure gives (many of) us a belief that there exists an artificial neural network (ANN) abstraction that is at least as capable as the human brain. Most commonly used ANN's, such as LSTM, have given us impressive performance in a number of domains, with works such as \citep{sutskever2014sequence,oord2016pixel,espeholt2018impala,berner2019dota,vinyals2019grandmaster}. The ANN's update their activations according to operations parameterized by weights, while they update their weights by taking small steps of stochastic gradient descent on some locally computed objective. However,  such weight update is likely too restrictive. Its clearest shortcoming is an inability to quickly store online information in weights, which therefore has to be stored in activations. However, utilization of a general weight update could be important for other purposes. Lack of this ability might make classic neural networks insufficient abstractions for capturing brains abilities on comparable size hardware. While a number of works discussed below aim to address this issue for classic neural networks, this begs the question: Can we simply learn the neural weight and activation update rules directly? In subsection \ref{evolution} we will discuss some of the approaches taken in this direction so far.

In this paper we pursue the idea that the activations and weights are updated based on local information (for example pre and post synaptic activations for a given synapse and synapse's previous state) and we conjecture that this update is likely parameterizable by a small number of parameters. Can we learn these parameters? In previous works, evolution as well as long term back-propagation training for the objective have been applied to address this problem. In this paper, we aim to provide a novel parameterization of the update rules as well as motivate and explore another means of training such networks that might prove to be more tractable, especially in longer lifetimes: That of training to remember the past. To make the discussion and relation to other works understandable, we start by explaining the parameterization of neural update rules that we use. Then, in section \ref{Motivation} we discuss how learning of the rules is relevant to many topics in machine learning, as well as discuss the relationship of this paper to previous works. In section \ref{FindingRules} we motivate the learning to remember objective. In section \ref{Training} we specify training protocol. In section \ref{FunctionalForms} we explain functional form details of the update rules. In section \ref{sec_experiments} we present experiments and finally we conclude in section \ref{sec_discussion}.

\section{Parameterization of local update rules.}
\label{Parameterization}

We model a neuron by the state (activation) of its body and the states of its synapses. As the real neurons and synapses are complex objects, we represent these states by small vectors, rather than scalars. We aim to parameterize the update rules that are applied at each computational step. The rules stated here are to different extents related to those proposed in other works, as we review in subsection \ref{evolution}.

Consider a neuron $i$ that receives information from $n$ other neurons $j=1,\ldots,n$. Let $h_i$ and $h_j$ be the activations of neurons $i$, $j$ and $w_{i,j}$ be the weights between neuron $i$ and $j$ (the state of the synapse). The variables $h_i, h_j, w_{ij}$ are small \underline{vectors}.

The state of the neural network consists of all the neural activations $h^t_i$, $i=1,\ldots,n$ and states of the connections $w^t_{ij}$, $i,j=1,\ldots,n$ where $n$ is the total number of neurons. The inputs and outputs are represented by special neurons.

At a given point in time $t$ the network updates its internal state according to some function: 
\begin{equation}
h^t_1,\ldots,w^t_{nn} = F(h^{t-1}_1,\ldots,w^{t-1}_{nn})
\label{eq:full_update}
\end{equation}
This is a function of many variables. We decompose it by enforcing locality: A given $h_i$ update can only depend on the activations of the other neurons through the synapses that connect to those neurons and on the neuron's previous state. A given synapse can only be updated based on the activations of the neurons it is connecting and the synapse's previous internal state. 

There are many types of neurons in the brain and thus it is reasonable to have many types of artificial neurons as well. We could use completely separate update functions for each neuron type. However, during evolution/learning of the update rules, a discovery made for one type of the neuron could not translate to the other neurons. Different types of the neurons in the brain share much of the genetic code for making them. For our neurons, we propose to represent such sharing by using the same neural network for each neuron, but to represent the different neuron types by different learned embedding vectors, that control the update of the update-rules-neural networks. We denote by $\tau_i$ the type of the neuron, and by $e(\tau_i)$ the embedding vector for a given neuron. The following are the full set of update equations.
\begin{eqnarray}
    u^t_{ij}, w^t_{ij} &=& g(w^{t-1}_{ij}, h^{t-1}_i, h^{t-1}_j, e(\tau_i), e(\tau_j)) 
    \label{eq:main_1} \\
    u^t_i &=& I(u^t_{i,1},\ldots,u^t_{1,j}, e(\tau_i))
    \label{eq:main_2} \\
    h^t_i &=& f(h^{t-1}_i, u^t_i, e(\tau_i))
    \label{eq:main_3}
\end{eqnarray}
Let us discuss these equations. The first equation is the computation of the synapse. The synapse takes its state $w^{t-1}_{ij}$, and the activations of the pre-synaptic and post synaptic neurons $h^{t-1}_j$, $h^{t-1}_i$, and computes new state $w^t_{ij}$ as well as the effect $u^t_{ij}$ on the post-synaptic neuron used next.

The next equation integrates the effect of all the synapses of the neurons. We use $u^t_i = \sum_j u^t_{ij}$. We could consider using a more complex rule, such as taking powers of $u_{ij}$ and a power of the resulting sum. However, this rule can be implemented by modifying functions $f, g$. We use the summation for the rest of the paper.

Finally, the last equation takes the integrated effect of all the synapses and the previous state of the neuron and computes the new state of the neuron.

In the implementation below we consider a system consisting of $N$ neuron types. The neurons are updated in sequence: All the neurons and synapses of a given type are updated first and the resulting values of these neurons are used for the next type of neuron. One could think of these as layers. However often times, a given neural layer (such as cortical layer) consists of several types of neurons, and the layer itself is repeated. Once a good way of finding the update rules is found, such architectures can be implemented and searched over.

\section{What could finding update rules do for us? Relation to other works.}
\label{Motivation}

\subsection{Episodic memory}

Remembering information even a few steps back has been a problem for early instantiations of neural networks due to the vanishing gradients \citep{hochreiter2001gradient}. This problem has been addressed by long short term memory (LSTM) \citep{hochreiter1997long} - an architecture containing gates that protects the information in activations. This extended the ability to remember to lengths of perhaps hundreds of steps, depending on the problem. However, to go beoynd this window, to store larger amount of information than what can be easily put into activations, and to have an access to the past that does not depend on how for something happened, a number of architectures primarily based on attention mechanism (but not only) has been proposed \citep{bahdanau2014neural,weston2014memory,hung2019optimizing,graves2014neural,vaswani2017attention,dai2019transformer}. In most of these cases, the information is stored in slots, which can be thought as a special type of fast weight: Each time we write, we allocate a special neuron, and fully change its weights (for example writing key into incoming and value into out-coming weights, related to \cite{hinton2019transf}, private communication). Some architectures such as \citep{wu2018kanerva} aim to store in a compressed way, by updating many slots at once and already use matrices with fast writes for storage. 

Another set of works considers updates to weights that have fast components, \citep{ba2016using,miconi2020backpropamine}. In the latter they consider weights of the following form: $w_{ij}=w^{\mbox{slow}}_{ij} + \alpha_{ij} \mbox{hebb}_{ij}$. Here the Hebbian part is updated in a prescribed way using product of pre and post synaptic activations while $w$'s and $\alpha$'s are trained using standard back-propagation.

\subsection{Meta-Learning}

Another reason that a rule different from back-propagation might be appropriate is the following: The goal of agents is to perform well in the future, not on the data from the past. Agents discover new things through their life (or we might want to apply networks to situations that are different from the ones we trained on so far). To find how to do this, we can learn how to adapt/learn - which is often referred to as meta-learning. The setting often used is one where the agent has an endless amount of lifetimes in changing environments. From this experience it needs to learn how to adapt when faced with a new changing environment (changing environment is often implemented as switching tasks or data). To solve this problem we in principle don't need anything special other than long enough back-propagation: If the changes of environment are within back-propagation period, the recurrent network can learn how to adapt to the changes. This approach was pursued in \citep{duan2016rl,wang2016learning}. In this approach, if using a standard recurrent network such as LSTM, one is looking for one (fixed) set of weights, so that the activations could learn how to handle the changing environment. That means that all the adaptation would have to happen in the activations and not weights. That is, the agent would not be, what we often consider to be learning (changing weights) when faced with a new elements of an environment.

One step towards changing the weights during new experiences is to store these new experiences in slots, which can be thought of as fast weights, and make predictions based on the stored data using an attention mechanism - to use episodic memory. This was pursued in \citep{santoro2016one,vinyals2016matching}, the latter in the context of classification, where few examples of novel classes were presented, stored, and then attended over when new examples to be classified were presented.

One model that does weight updates to the main network when adapting to the changing environment is model agnostic meta learning \citep{finn2017model}. It updates the weights using the standard gradient descent - taking the steps on the new examples. However, it aims to find the initial good weights, in such a way that subsequent steps in the new tasks would lead to a quick adaptation. Such weights are found by directly optimizing the novel tasks objective, which requires back-propagation through back-propagation, and constitutes a rather complex way of updating the weights.

A simpler weight update rule for achieving this purpose was proposed in \citep{nichol2018first}. The network contains slow and fast weights. Given a new task, several steps of gradient descent on the fast weights are taken. Subsequently, the slow weights are updated by taking a small step in the direction towards the fast weights. 

The ultimate dream (not realized yet) of the approach described in this paper and related works is to figure out general weight and activation update rules that would (meta) learn and let the network decide how these weights should be used, whether similarly to the effect of classic back-propagation, or fast learning for storage, or the appropriate updates for adaptiation to new tasks using both activation and weight updates, or potentially other uses. For example the algorithm of the previous paragraph can be represented by making one component of the weight state being the slow weight and another component being the fast one. While achieving such general updates sounds like a difficult task, we reason that all we essentially need, as far as neural network is concerned, is to learn two, not so complex functions.

A step in this direction is taken in \citep{ravi2016optimization}. They again consider meta-learning classification task - learning so that when presented with examples of new classes, the network learns to classify them quickly. Rather then taking a gradient on the new examples they feed this gradient (and other information) into weight update rules, that are parameterized by an LSTM, letting the LSTM decide how or how much to update a given weight on the new examples presented. Because of the nature of the problem - classifying the new examples - only a short back-propagation (over these examples) is needed to train the weight update rules. In our paper we are looking for general update rules, that work for long term and do not use gradient information.

\subsection{Objectives}

Standard neural networks compute gradients of some objective. In a (image) classification problem the loss seems relatively straightforward - a measure of an error of the prediction of label given input. However, even in this case, this might not be the best objective - because we might want the network to extrapolate out of distribution (or if we are in setting of having unlabeled examples), we might want a different form of training, such as meta-learning, or to impose a structure understanding loss such as generative model \citep{oord2018representation,donahue2019large,jaderberg2016reinforcement,gregor2019shaping}. The situation is even more complex in reinforcement learning, where the agent is not given the right targets (actions), it needs to explore, it determines its own data and where it is less clear what a good objective (reward) is. One approach \citep{xu2018meta,veeriah2019discovery} is to meta-learn an auxiliary loss function, such that its gradients implement a good RL algorithm. These gradients are a special way to produce weight updates, and we could possibly instead learn such updates directly.

\subsection{Neuro-evolution and meta-learning neural networks.}
\label{evolution}

Rather than designing neural networks architectures and the update rules by hand, there have been a number of works that attempt to find them automatically \citep{soltoggio2018born,stanley2019designing}. There are two common ways to perform such search - evolution and (meta) gradients. In the former one keeps a population of neural networks and, one way or other, mutates them, computes fitness of each and performs natural selection. In the latter, one consider an entire experience or a long part of it and performs back-propagation.

Evolution can be applied to various aspects of neural networks. One type of application is to evolve the structure of a neural network, such as number of layers, connectivity and non-linearities while using the standard back-propagation algorithm to update the weights \citep{bayer2009evolving, rawal2018nodes, real2019regularized}. Another approach is to take a given architecture and evolve the values of weights that solve a task \citep{wierstra2008natural,salimans2017evolution}.

There are a number of works that aim to learn local neural network update rules (as done in our paper). In one of the earliest works, \citep{bengio1992optimization}, they consider finding an update rule of a (scalar) weight as a function of local information in the network: A linear combination of local products and non-linearites of pre-synaptic and post-synaptic neural values, the previous values of weights and an overall modulatory signal. In HyperNEAT \citep{risi2010indirectly}, compositional pattern producing network (CPPN) is used to produce weights of the actual network. Neurons are laid out on a grid and the CPPN takes their coordinates and produces the weight value. The weights of the CPPN are evolved. In the adaptive version, pre-synaptic and post-synaptic activations, as well previous neural weights are passed to the CPPN as well, making the CPPN essentially implement (position dependent) local learning update rule. There are a number of other works in similar directions \citep{schmidhuber1992learning,vassiliades2013toward,orchard2016evolution,miconi2016learning,miconi2018differentiable,gu2019meta,munkhdalai2019metalearned}.

The idea of representing activation and synapses by vectors and using general networks for their update was very recently, and independently proposed by \citep{bertens2019network}. As an application they evolve a system of seven such neurons to solve a T-maze task. The differences in our work are the following: We train the network for next step prediction (language) model, motivate and investigate a different objective - that of remembering, (meta) train by back-propagation, study the computational abilities of these networks as a function of problem difficulty and various parameters, and find somewhat different update rules that work well.

Another hint why a vector representation is needed, is the existence of an online algorithm for calculating recurrent network gradients - real time recurrent learning (RTRL) \citep{williams1989learning}. The RTRL does not need to go back in time to calculate gradients. However it needs to store a non-local and an un-tractably large number of parameters online: derivatives of a given neuron with respect to all other weights ($\partial h_i/\partial w_{jk}$). One can consider a truncation of such rule that would only need much fewer parameters. Finding the weight update rules directly might effectively and in particular find a good truncation of such information.

A common theme for number of these works (including ours) is that one set of parameters (for the update rules) - meta-parameters - parameterizes another set of parameters (weights and activations). 
These networks can be thought of as standard RNNs with unusual functional form where activations and weights together form what are normally activations (state) of an RNN and the meta parameters what are normally weights. The primary difference from the standard RNN's is that we are looking for a small set of meta-parameters, which means the algorithm needs to \emph{learn to store} all information about the data in the network's state (weights and activations) rather then in the meta-parameters - see the detailed discussion on this in subsection \ref{subsec_learn_to_remember} on back-propagation and long term training, and the discussion in section \ref{sec_experiments} on what it means for something to be a learning rule. However, one can relax the assumption of smallness and obtain a well performing system trained by standard back-propagation, as done in Hypernetworks \citep{ha2016hypernetworks}. There, they use a set of parameters (hyper-weights) that parameterizes the weights of convolutional networks and LSTM. However, it still contains on the order of 100K parameters (hyper-weights), which means it is still sensible to train this system by direct, short term back-propagation for the objective, utilizing the meta-parameters to store the information about the data.

\subsection{AI generating algorithm}

In \citep{clune2019ai} it was proposed that finding a general AI could proceed by meta-learning/evolving three pillars: 1) Structures of neural networks, 2) Learning algorithms 3) Environments/training data. Structures of neural networks are properties such as connectivity of layers, sizes, non-linearities and weight initializations. Learning algorithms are the ways the internal variables are updated during agent's interactions with the environment. Meta-learning/evolving the environments is a process by which more and more complex and diverse problems are produced for or by the agents. We propose that learning the parameters of two relatively simple functions proposed here might be a sufficient substrate for the second pillar.

\section{Finding the rules}
\label{FindingRules}

We need to find the functions $f, g$ in (\ref{eq:main_1}, \ref{eq:main_3}) that solve complex problems we are interested in, such as sequence modeling or reinforcement learning. There are two basic ways for training such rules directly for the objective we are interested in - evolution and long-term back-propagation. We discuss what are their advantages and disadvantages and motivate the objective used in this paper, relating it to back-propagation training. 

\subsection{Evolution}

We could evolve the parameters of the functions $f, g$ in (\ref{eq:main_1}, \ref{eq:main_3}) to solve complex tasks or survive in an environment. Under standard settings, a population is created, run for their entire lifetime, and then natural selection is applied. The advantage of this approach is that it directly trains for what we are interested in, and the search is (often) non-local. However as the problems get more complex, this step of the evolution loop becomes longer and longer. It might be hard to be able to make enough evolutionary steps to find the right algorithm. There can be ways around this problem, such as not resetting the learned weights from scratch at every iteration \citep{jaderberg2017population}, and we certainly consider this to be an exciting direction of research.

\subsection{Learning to remember}
\label{subsec_learn_to_remember}

Let us ask a question - why can't we train the function $f, g$ in the standard way - run the network forward $T$ steps, where $T$ is typically of the order of hundred, back-propagate and update $f, g$? 

Consider training a next step prediction language model. First let us see what happens when we train a standard network such as LSTM. We consider the situation where we read the whole text in the order it is written. At some time $t$, the state of the LSTM is described by $h^t,c^t,W^t$ where $W$'s are the weights of the LSTM. Next, we take the next $T \sim 100$ inputs, keep $W$'s fixed, and update $h,c$'s $T$ times to arrive at state $h^{t+T},c^{t+T},W^t$. Then we back-propagate and update the $W^t$ using a gradient step to obtain $W^{t+T}$. This will increase the likelihood of the sequence we have just seen. What we are really interested though is to increase the likelihood of the sequences that we will see in the future. Why does this work?

A given snippet of text typically contains a structure within the back-propagation window. For instance, the letters "dee" are usually followed by the letter "p".
This structure might not appear again for thousands of steps. However, each time such structure is presented, the algorithm effectively \emph{memorizes} a piece of this information because it is taking small steps of the gradient. It is \emph{not} memorizing \emph{because} it needs this information later, it is memorizing it because that's how it is built - to slowly memorize (in a compressed fashion). The trouble arises when the structure spans much more then the back-propagation period. If an event $B$ (harvesting crops) follows an event $A$ (such as planting seeds) thousands of steps later, the connection $A \rightarrow B$ is not found - it is not put into memory through gradient (or only accidentally). Our standard back-propagation algorithm is inadequate, and modifications needs to be found. Solutions to these problems are memory architectures.

In our case, we are only learning by back-propagation the small number of (meta) parameters that control how weights and activations are stirred in response to inputs. In fact we are looking for a single set of parameters of the update rules, so that starting from the beginning of training, the network learns to be good at the task. If training end to end (say for the next step prediction) any information that is present now needs to be stored \emph{only because} it is useful later. For example the network needs to remember that $p$ follows $dee$ only because it will need this information thousands of the steps later. The problem that faced the standard neural networks only when long dependencies are considered, manifests itself here right from the start. But perhaps we could consider this to be an opportunity - figure out one principle that interpolates both cases.

There are essentially two solutions to this problem. One is to push hard the end to end approach. Figure out how to optimize the long term objective, either through evolution, long term back-propagation or some other yet not invented trick.

The other approach is to push the memorization all the way. Rather then taking a small step of parameters in the direction of gradient, we aim to memorize everything. This should include one shot memorization of what just happened and a long term memorization of the statistics of the data, with some trade-off given by the capacity of the network and the objective we impose for this purpose. Such memorization is sometimes pursued by various memory architectures mentioned, which typically split the weights into slow weights trained by back-propagation and fast weights (such as memory slots) that are updated quickly in a particular, perhaps hand-designed way, for example by storing input or hidden state at every time point. In our approach, the hope is that the networks learn which weights should update slowly, which fast and which perhaps at an intermediate speed, acting for example as a working memory with higher capacity and perhaps easier to write to than what activations could provide.

Even in this approach, there are many ways one can imagine imposing a memorization objective. We propose one simple way, but this is very much an open problem. Finally, there are a few points we would like to make regarding potential advantages of this method.

a) It has been shown empirically that the networks that are trained long enough to produce zero training errors actually generalize well \citep{feldman2019does}. One explanation of this phenomenon is long tails. In a situation where a rare example arrives, if the network has memorized a previous similar rare example, it can make correct prediction.

b) Because of a limited capacity, the network trained to memorize might learn to compress the information, unlike many memory architectures which store all latent vectors in slots. Compressive models might be more efficient storage for search and predictions.

c) Finally the most basic approach to meta-learning is simply a long enough back-propagation period. However, as discussed in the introduction, such method can only learn to adapt to changes using neural activations. Instead, several works approach meta-learning by storing the past data, which is what our objective does, and use this storage to make predictions.

\section{Data and training procedure}
\label{Training}

The problem we are going to study is a next step prediction in a sequence. As the data we take the text of the "Wizard of Oz", a book with approximately 232k characters. To keep the input dimension low, we convert all the characters to lower case and group all the non-letter characters into a single symbol. The resulting input is an one-hot vector of size 27.

We use the following training process. We read character by character, wrapping around once the end of the text is reached. Every $T=10$ steps, we have a sequence $x_{t-T+1},\ldots,x_t$ of the last $T$ characters. We train to remember past sequences of length $T$.

For the past, we pick a random time $t_p$ uniformly from an interval $(t-P, t-T)$ where $P$ determines how much past we are aiming to remember. We set the initial state of the network $(h_p, w_p)$ by taking the most recent weights and resetting activations to zeros: $(h_p, w_p) = (0, w_t)$. We run the network on the sequence $x_p,\ldots,x_{p+T-1}$ and compute the prediction loss $L_p$ (the $h$ is only reset for the past run and continues online without resetting). We train the meta parameters by back-propagation through the past and the current sequence (through $2T$ steps). 

\section{Neural updates functional forms}
\label{FunctionalForms}

Here we discuss the functions $f, g$ that we use in equations (\ref{eq:main_1}, \ref{eq:main_3}). We use a number of functional forms that update the state of $h$ and of $w$ given their inputs. We drop the time index and instead use the upper index to denote components of the state. As mentioned, the states $h_i$ and $w_{ij}$ are small vectors (with sizes between 2-14), with $h^\alpha_i$, $w^\alpha_{ij}$ denoting $\alpha$-th component of these vectors (a scalar). We use the synaptic update $f$ of the form
\begin{eqnarray}
    \label{eq_uij}
    u_{ij}^\alpha &=& w^\alpha_{ij} h^0_j \\
    \label{eq_xij}
    x_{ij} &=& [h_i, h_j, u_{ij}, h_i h^1_j, \tau_i, \tau_j] \\
    \label{eq_w}
    w_{ij} &=& OP(w_{ij}, x_{ij})
\end{eqnarray}
Where the $OP$ is one of the operators is described in the next paragraph. We write the integration step, and the activation update $g$ as
\begin{eqnarray}
    u_i &=& \sum_j u_{ij} \\
    h_i &=& OP(h_i, [u_i, \tau_i])
    \label{eq_h}
\end{eqnarray}
For the $OP$ we use the following list of operations. 1) MLP: One hidden layer feed-foward neural network. For the $h$ operations sigmoid non-linearity is applied at the output, while for the $w$ update, no non-linearity is applied, or, denoting MLP-tanh, tanh non-linearity is applied. The hidden layer non-linearity of the MLP is tanh. 2) LSTM: Standard LSTM update is used where the state ($h$ or $w$) is split into hidden and cells of the LSTM. We consider few variations. In the standard LSTM, the final operation is $h = g_o \tanh(c)$ where $c$ is the cell and $g_o$ is the output gate. We denote LSTM-$\sigma$ when using $h = g_o\sigma(c)$ and LSTM-id when using $h=g_o c$. Finally, we also use the following operation for OP($s$, $x$) (with $s$ either $h_i$ or $w_{ij}$), which we denote by GATED:
\begin{eqnarray}
    a, b &=& MLP([s, x]) \\
    g &=& \sigma (a) \\
    r &=& \tanh (b) \\
    s_{new} &=& (1-g) s + g (r + h_i h^1_j)
    \label{eq_gated}
\end{eqnarray}
The overall architecture of the system consists of one or more hidden layers with the updates just described and an input layer. All layers are connected to all other layers including themselves. The operations of the input layer are as follows, with $y$ denoting the input to be predicted.
\begin{eqnarray}
    v_i &=& MLP([h_i, u_i, \tau_i])) \\
    h^\alpha_i &=& \frac{\exp(v^\alpha_i)}{\sum_k \exp(v^\alpha_k)} \\
    p_i &=& h^0_i \\
    L &=& -\sum_i y_i \log(p_i) \\
    h_i &=& (y_i, h^1_i,\ldots,h^{n-1}_i, y_i-p_i)
\end{eqnarray}
The second equation is the softmax function. In the third equation, the first component of $h$ is used as the probability. In the fourth equation, the standard cross-entropy loss is calculated. In the final equation we replace the first component of $h_i$ by $y_i$ and the last component by the prediction error $y_i-p_i$. The weights $w$ were initialize at random at the beginning of training in the same way as if they were weights of standard neural network.

\section{Experiments}
\label{sec_experiments}

We made a number of experiments before we arrived at the functional forms stated above. We describe some observations we made during that time. First, note that the effect of the synapse on the neuron, the $u_{ij}$ in (\ref{eq_uij}), has a particularly simple form of multiplication. This, together with the summation over $j$ becomes the standard matrix multiplication used in neural networks, except that there are essentially neuron state size number of such operations (one for each $\alpha$ in (\ref{eq_uij})). We experimented with merely using the OP in (\ref{eq_w}) to output this value, but we found that fails to learn well. Also note that we used 0-th component of $h$ in (\ref{eq_uij}) and the first component in equations (\ref{eq_xij}, \ref{eq_gated}). We found this combination (using different components) to work well, but undoubtedly, significantly more exploration can be done in this direction. 

\begin{figure*}[ht!]
\vskip 0.2in
\begin{center}
\centerline{\includegraphics[width=18cm]{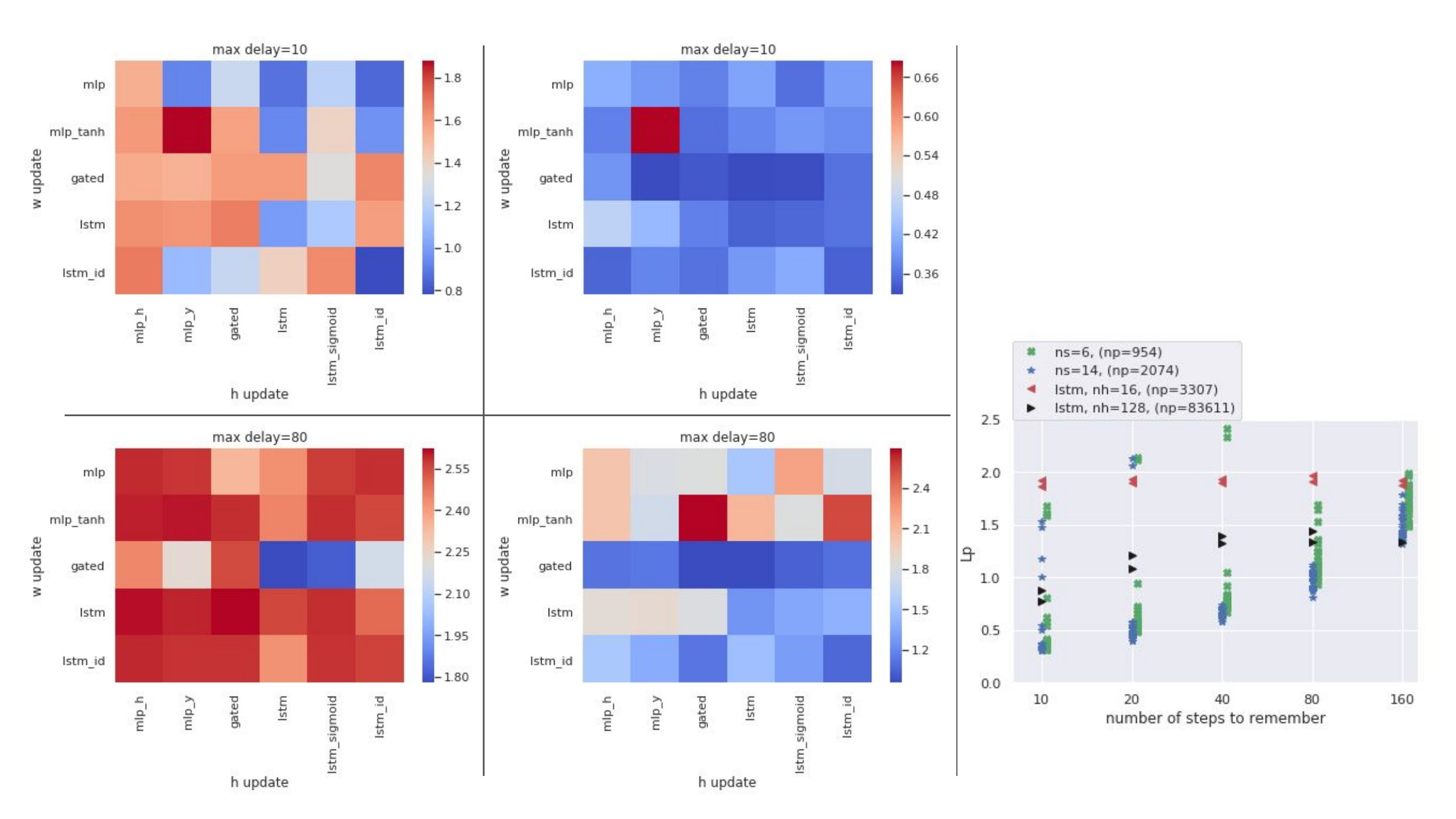}}  
\caption{{\bf Left}: Average loss early in the training for delays of 10 (top) and 80 (bottom) for different update rules of activations and weights. {\bf Middle}: The same but at the end of the training. See text for details and the discussion. {\bf Right}: Blue and Green: Performance for different delays and two states sizes (6 and 14). Each dot is one run. Red and Black: Standard LSTM with 16 and 128 units. The parameter $np$ shows number of parameters that are trained by back-propagation.}
\label{fig_experiments}
\end{center}
\vskip -0.2in
\end{figure*}

Now we turn to the results based on the equations stated in the previous section. In Figure \ref{fig_experiments}, left and middle, we analyze the effect of different OP's in both $h$ and $w$. We analyze the performance on delays of 10 and 80 (10 means up to $T(=10)+P(=10) = 20$ steps into the past from the most recently seen data-point). We are interested in the performance both early in the training and also once a lot of the curves have already converged. To get the early performance, we took the average of losses over the first 200k steps. This number was obtained roughly at the location after some of the curves dipped significantly. The final performance was taken at 1M time steps. For a given setting we made a number of runs. Out of this both the average curves and the best curves are interesting. Average means how does a given system performs on average. The best performance shows how well a given system can perform if the optimization succeeded more often (more about what is the potential of the functional form). We display the plots for the best performance, but the plots of the average performance give essentially the same conclusions. 

The middle column Figure \ref{fig_experiments} shows the final performance. For short delays, essentially all combinations succeed in reaching good performance (except MLP-tanh x MLP-y). However for longer delays, there is a significant difference between the runs. What emerges is that gated network for the weight update consistently performs the best. However, for certain updates to $h$, LSTM-id performs well. MLP networks for weight updates generally fail. The reason for this probably is similar to why standard recurrent neural network doesn't learn well compared to LSTM - the weights need a good mechanism for remembering. For example, in standard gradient descent, the update is $w_{ij} \leftarrow w_{ij} - \eta \nabla L$. With the gated networks, the network could learn to implement close to this additive update for some neuron types, and fast updates for others, or it can learn to make fast update depending on the current input.

The left column of Figure \ref{fig_experiments} shows the performance early in training. This time, for longer delays, LSTM update on $h$ produces stronger performance. 

Figure \ref{fig_experiments} right, shows performance of several runs for different settings. We consider state sizes of $6$ and $14$ (equal for both activation and weight states). We also run standard LSTM baseline on this task. We make a number of observations from this figure. First we see that there is quite a spread of the values that different runs of the same hyper-parameter produce. Thus, there is clearly an optimization problem. Second, the larger state size ($14$ compared to $6$) clearly works better. Third, we see that for delays up to 80, our system works significantly better than LSTM's displayed.

It is interesting to compare our network to standard LSTM. Our network can be thought of as a classic neural network with an unusual functional form, where the meta-parameters are the parameters trained by back-propagation and the activations and weights are what would normally be activations of the recurrent network - the recurrent state. In this interpretation, the size of the state in our network is significantly bigger than that of LSTM, while the number of parameters that control these activations is quite a bit smaller - this number is displayed in the Figure under variable $np$. In our network we have a relatively small number of variables controlling a system with a very large state size.

The property of finding a small number of weights that controls a much larger state space has been encountered before in Hypernetwork \citep{ha2016hypernetworks} (where the ratio is not very large) and in non temporal setting of de-noising auto-encoder \citep{fernando2016convolution}. As this ratio gets smaller and smaller, the more the network updates have to resemble a learning algorithm - the more of the learning has to be learned. And, as we discussed in sub-section \ref{subsec_learn_to_remember}, the less it is possible to learn this using direct back-propagation on short sequences, which is what motivated the learning to remember objective of this paper. 
Have we succeeded at what we would call learning rules? And what does it mean for something to be a learning rule, as opposed to just a network of small number of parameters, parameterizing a large state. First, a given learning rule should work rather generally - that is a given fixed set of meta-parameters should be able to solve many problems. This would likely require producing a diversity of training, validation and testing problems and we hope such analysis will be done in the near future.

There is another, simpler property though that a learning rule should satisfy. In our networks, a given set of meta-parameters is independent of the number of neurons. A good learning rule should be able to utilize these neurons, that is, the performance should increase as we increase the network size. 
Unfortunately, we haven't found this to be the case in our system. We found that even with small number of neurons we get the above performance. There could be a number of reasons for this. First, the layer containing the inputs has self connections, and a lot of the computation the network needs to do could be done there. The hidden layer does have an effect especially early in the training (as seen in Figure \ref{fig_experiments} left), but it goes away more later on, possibly with the network learning to primarily utilize self connections. Second, we have seen that the optimization is a problem and it could be causing large network not to learn as well as they could. 

While this scaling has not succeeded so far, there is significantly more exploration that can be done in this direction. We hope we have provided a interesting parameterization, analysis and insights into this problem that will be prove useful in the future. 


\section{Discussion and conclusions}
\label{sec_discussion}

In this paper we were motivated by a tantalizing possibility that all the learning algorithms that we need, from one shot learning, model learning, reinforcement learning could potentially be parameterized by two simple functions: The update rule for activations and the update rule for weights, with different neuron types represented by embedings that affect the update rules. We explored a tiny corner of this space - hand-searched for the update rules, learned to remember and applied to text. Just like rectified linear units made a big difference for training deep networks, there are likely to be many innovations that could make this approach much better.

For the future, research can be done in several directions. First, we could perform evolutionary search both for the architectures of the update rules as well as the network architectures. Then, different training regimes and different approaches to optimizing these rules could be tried. Finally, the problems on which these networks are applied are likely just as important as the architectures and update rules \citep{clune2019ai}.

Can a general learning algorithm be represented by such simple functions and an evolutionary search? The answer, we believe, is that this algorithm does not stand alone. As agents move through life, they not only acquire new experience but also improve the way they learn. A good learning algorithm is determined not only by the correct settings of the update rule parameters, but also by the values of the actual weights of the network. While we are born without much specific knowledge about the world, some of the initializations of these weights, as well as environmental settings, could prove important for starting the network at the right place at the beginning of agent's life. We could aim to evolve/meta-learn such initializations. Alternatively, since we don't have a restriction of finite life on our agents, we could simply continue its life forever, or find some other mechanism for starting a new agent. 

\bibliography{references}
\bibliographystyle{icml2020}

\end{document}